# Implementing a Bayesian Scheme for Revising Belief Commitments

*Lashon B. Booker, Naveen Hota†, and Gavin Hemphill‡*

Navy Center for Applied Research in AI
Code 5510, Naval Research Laboratory
Washington, D.C. 20375

*ABSTRACT*

Our previous work on classifying complex ship images [1,2] has evolved into an effort to develop software tools for building and solving generic classification problems. Managing the uncertainty associated with feature data and other evidence is an important issue in this endeavor. Pearl [7-11] has developed a Bayesian framework for managing uncertainty that has proven to be applicable to several of the belief maintenance functions that are necessary for classification problem solving. One such function is to determine a *belief commitment* which designates in categorical terms the most probable instantiation of all hypothesis variables given the evidence available. Before these belief commitments can be computed, the straightforward implementation of Pearl's procedure involves finding an analytical solution to some often very difficult optimization problems. We describe an implementation of this procedure using tensor products that solves these problems enumeratively and avoids the need for case by case analysis. The procedure is thereby made more practical to use in the general case.

## Introduction

A prototype system for classifying complex ship images has convincingly demonstrated that Bayesian reasoning is a valuable tool for making plausible inferences about classificatory hypotheses given impoverished feature data [1,2]. It remains to be shown that such methods are also useful in handling the large scale, real-time and resource-constrained classification problems that are of interest to the Navy. Real-time classification in an operational environment is a demanding task. Regardless of the kind of sensor information available — visual, infrared, radar, or sonar — this is a task in which complex inferences must be made reliably under stringent computational constraints, and based on incomplete and uncertain evidence. Our current research efforts are focused on devising a robust and accurate classification problem solver that meets this challenge.

Managing the uncertainty associated with feature data and other evidence is an important aspect of these classification problems. The Navy problem domain is one in which the reliability of machine-drawn inferences is a critically important issue. It is therefore crucial that evidential uncertainty be managed with representations and rules of combination based on sound theory and clear semantics. We have determined that a Bayesian approach is the best choice for the problems of interest here [1]. Furthermore, among the many Bayesian schemes available, the approach championed by Pearl [7-11] has the required breadth, flexibility, and conceptual clarity [2].

Pearl's framework includes procedures that happen to be appropriate for several of the belief maintenance functions associated with classification problem solving. One of these functions is to determine a *belief commitment* which is an optimal, categorical instantiation of every hypothesis variable based on the evidence available. Pearl's algorithm for revising belief commitments is a distributed computation that solves a global optimization problem — finding the most probable instantiation of all variables — by recursively computing local maximizations. The method has been shown to be effective in a diagnostic task [5], and is clearly applicable to the classification problems discussed here.

One practical difficulty, though, is that in every reported implementation of the procedure the local contributions to the global optimization problem were solved analytically. A generic classification problem solver of the kind envisioned here cannot count on being provided with handcrafted analytical expressions for solving every new problem. Accordingly, we have devised an efficient, enumerative implementation of Pearl's procedure that avoids this difficulty.

This paper begins with a discussion of generic classification problem solving and the importance of belief commitments in that process. In the subsequent sections we review Pearl's procedure for revising belief commitments, and then describe the way we have implemented that procedure.

**Classification Problem Solving**

Classification problem solving is a ubiquitous activity in knowledge-based systems [4]. The classification problems found in the unstructured environments of the Navy domain are especially difficult and challenging. These problems involve working with data at several levels of granularity, ranging from raw sensor returns to partially reliable intelligence reports about object identification. Several methodologies are available for transforming this data into evidence useful for classification: signal processing, pattern recognition and feature extraction algorithms, and knowledge-based approaches to feature interpretation. The goal of classification problem solving is to achieve a coherent analysis of sensor returns by selectively applying the methodologies to the data, then choosing the most plausible solution. Orchestrating this entire process is a complex job. Our approach to managing this complexity is to identify the information processing requirements common to several Navy classification problems with data derived from several kinds of sensors. Such an understanding will provide insights about the kinds of knowledge structures and control regimes that are characteristic of these classification tasks in general. This, in turn, will allow us to develop system building tools that reflect the inherent structure of the classification problems, facilitating system design, knowledge acquisition, and explanation. An added benefit is that these tools can then be used for a wide variety of applications.

This approach builds on work done by Clancey[4] and Chandrasekaran [3], who argue that the inherent structure of any problem solving task can be revealed by decomposing it into elementary organizational and information processing strategies called *generic tasks*. Because many heuristic programs accomplish some form of classification, a considerable amount of work has already been done to analyze classification problem solving in this way.

Perhaps the most important issue emerging from a generic task analysis of classification problem solving is the management of uncertainty. Uncertainty has many potential sources in Navy problems: sensors are rarely completely accurate or reliable; observations and feature extraction techniques can be flawed; there may be no strong correlations between manifestations and causes; some of the evidence may be contradictory; and so on. The reasoning mechanism will therefore have to perform causal, diagnostic and intercausal inferences in combination. Effective interaction with feature extraction modules will involve decisions about the order to acquire data, the number of image frames to process before making a judgement about some feature, etc. Belief updating therefore must be amenable to a variety of control schemes. We have determined that a Bayesian approach to belief updating is the best choice for our classification problems [2]. Furthermore, among the many Bayesian schemes available, the approach championed by Pearl [7-11] has the required breadth, flexibility and conceptual clarity.

Pearl's framework provides a method for hierarchical probabilistic reasoning in directed, acyclic graphs called *belief networks*. Each node in the network represents a discrete-valued propositional variable. Each link between nodes represents a causal dependence or object-property relationship whose strength is quantified using a matrix of probabilities conditioned on the states of the causal variable. The directionality of the links is from causes to manifestations, each link emanating from a parent node in the graph. It is important to note that numbers used to quantify the links do not have to be probabilities. All that is required is that the matrix entries are correct relative to each other.

The belief updating scheme keeps track of two sources of support for belief at each node: the diagnostic support derived from evidence gathered by descendants of the node and the causal support derived from evidence gathered by parents of the node. Diagnostic support provides the kind of information summarized in a likelihood ratio for binary variables. Causal support is the analogue of a prior probability, summarizing the background knowledge lending support to a belief. These two kinds of evidential support are combined to compute the belief at a node with a computation that generalizes the odds/likelihood version of Bayes Rule. Each source of support is summarized by a separate local parameter, which makes it possible to perform diagnostic and causal inferences at the same time. These two local parameters, together with the matrix of numbers quantifying the relationship between the node and its parents, are all that is required to update beliefs. Incoming evidence perturbates one or both of the support parameters for a node. This serves as an activation signal, causing belief at that node to be recomputed and support for neighboring nodes to be revised. The revised support is transmitted to the neighboring nodes, thereby propagating the impact of the evidence. Propagation continues until the network reaches equilibrium. The overall computation assigns a belief to each node that is consistent with probability theory. See Pearl [10] for more details.

While computing the impact of evidence on degree of belief in a hypothesis is crucial to a classification problem solver, it is not the only belief maintenance function required. Equally important is determining how the various degrees of belief in classificatory hypotheses are to be interpreted. In complex classification problems it is very important that the rationale for beliefs be explainable in easily understood terms. Generating a coherent explanation involves the simultaneous acceptance of a set of hypotheses, a requirement that goes beyond simply noting the degree of belief in any individual hypothesis. This means that the problem solver must make a *commitment* in categorical terms about the best way to instantiate each hypothesis variable based on the evidence available. Pearl [9,11] has recognized the need for computing belief commitments and shows how it can be done within a Bayesian context.

**Revising Belief Commitments**

In Pearl's framework, the problem of finding the most probable instantiation of all hypothesis variables given some body of evidence $e$ can be easily stated in terms of conditional probabilities. If $W$ is the set of all hypothesis variables being considered, then any assignment of values $W = w$ to all these variables is called an *interpretation* of the evidence $e$. The belief commitment we want is given by the interpretation $W = w^*$ which satisfies $P(w^* \mid e) = \max_{w} P(w \mid e)$. In other words, $w^*$ is the most likely interpretation of the evidence.

Finding $w^*$ is a very difficult optimization problem that Pearl makes tractable by using the independence relationships among variables, reflected in the network topology, to recursively decompose the global computation into local maximization problems. The best interpretation for any given variable can be determined by finding the best partial interpretation in independent subgraphs of the network. This decomposition can be applied recursively, halting at the network periphery when instantiated evidence variables and boundary conditions are encountered. The updating scheme itself is a parallel message passing process which makes use of two kinds of messages at each node $X$: the probability of the most likely interpretation of each of the node's ancestors in the subgraph above the link from that ancestor to $X$; and, the probability of the most likely interpretation of the node's descendants in the subgraph rooted at $X$ for each hypothetical instantiation of $X$. These messages, together with a matrix of conditional probabilities relating the node to its immediate ancestors, are all that is required to revise belief commitments. Incoming messages serve as an activation signal, causing the probability of the most likely interpretation for that node to be recomputed. Messages influencing the interpretation of neighboring nodes are revised and transmitted, thereby propagating the new information. Propagation continues until the network reaches equilibrium.

350

When a given node is activated, it combines the messages from its neighbors to determine its best value and update its outgoing messages. This part of the procedure can be summarized as follows. Using Pearl's notation, let $X$ be a node having $n$ parents $U_1, \cdots, U_n$ and $m$ children $Y_1, \cdots, Y_m$. $X$ will receive a message $\pi_X^*(u_i)$ from each parent and a message $\lambda_{Y_j}^*(x)$ from each child. With these messages and the set of fixed probabilities $P(x \mid u_1, \cdots, u_n)$, $X$ computes three quantities:

1) $BEL^*(x)$, the probability of the most probable interpretation of the evidence consistent with the assignment $X = x$, which is given by

$$BEL^*(x) = \alpha \max_{u_k, 1 \leq k \leq n} F(x, u_1, \cdots, u_n)$$

where $F(x, u_1, \cdots, u_n) = \prod_{j=1}^{m} \lambda_{Y_j}^*(x) P(x \mid u_1, \cdots, u_n) \prod_{i=1}^{n} \pi_X^*(u_i)$

Using $BEL^*(x)$, $X$ can directly determine its optimal value $x^* = \max_x^{-1} BEL^*(x)$.

2) A $\lambda^*$ message for each parent $U_i$ giving the probability of the most likely interpretation of the assignment $U_i = u_i$ in the subgraph below $U_i$ rooted at $X$. This is computed from

$$\lambda_X^*(u_i) = \max_{x, u_k : k \neq i} \frac{F(x, u_1, \cdots, u_n)}{\pi_X^*(u_i)}$$

3) A $\pi^*$ message for each child $Y_j$ giving the probability of the most likely interpretation of the assignment $X = x$ in the subgraph above $Y_j$ that includes $X$. This is computed from

$$\pi_{Y_j}^*(x) = \alpha \frac{BEL^*(x)}{\lambda_{Y_j}^*(x)}$$

More details can be found in [11].

The heart of this computation is the optimization problem that must be solved in the determination of $BEL^*$ and $\lambda^*$. Pearl gives two examples of how to do this, one in a network for medical diagnosis [11] and another in a network for diagnosing faulty digital circuits [5]. In each case, the first step in solving the problem is to find an analytical expression for the maximum of $F$ that can be easily computed. The form of this expression of course depends on the link probabilities quantifying the interactions among causes for a given symptom. In both diagnostic problems these probabilities have a canonical, straightforward definition that applies uniformly to all links in the network. Even so, maximizing $F$ proved to be a non-trivial exercise. The circuit diagnosis problem required an exhaustive case by case analysis in order to compute the maximum of $F$. The medical diagnosis problem did not lend itself to any analytically determined optimal solution, so the general case had to be handled with a heuristic greedy algorithm.

The difficulty of maximizing $F$ raises serious concerns about how practical the belief revision computation will be in an arbitrary belief network. To some extent these concerns can be met by following Pearl's suggestion [10] to use canonical models of typical interactions among causal conditions. Examples of such canonical models include disjunctive interactions (the "noisy OR"), conjunctive interactions (the "noisy AND"), and various enabling mechanisms controlling when an influence is effective. These models can be mathematically analyzed in advance and then used as required in any given network. More complex interactions can be specified with Boolean combinations of these models, which are also amenable to mathematical analysis. All that needs to be done for a particular application is to refer to the analysis of the appropriate model to determine how to maximize $F$.

In many real applications, however, the only basis for determining link probabilities is a statistical analysis of a body of experiential knowledge, a simulation of a complex process, or something else that does not easily lend itself to a straightforward interpretation in terms of pre-analyzed interactions. The best available description of an interaction in these circumstances is the set of link probabilities themselves. This is particularly important in the classification problems of interest here. Most of the link probabilities are derived from object/property descriptions, observed frequencies, or other kinds of empirical data. Analytical descriptions of the interactions these numbers quantify are very difficult to construct. If the belief revision computation is going to be generally applicable, it must be tractable in situations like this where $F$ must be maximized without consulting a pre-computed analysis.

**Tensor Product Computation**

We have devised an implementation of the belief revision procedure using tensor products that maximizes $F$ enumeratively and avoids the need for extensive mathematical analysis. The procedure is thereby made more practical to use in the general case. In this section we review the tensor operators needed to perform the computation, then give the details of how they are used in the belief revision process.

A *tensor* is a mathematical object that is a generalization of a vector to higher orders. The order of a tensor is the number of indices needed to specify an element. A vector is therefore a tensor of order one and a matrix is a tensor of order two. Three standard operations defined on tensors are relevant to this discussion:

**Term Product**  The term product is defined between two tensors $\vec{A}$ and $\vec{B}$ having the same indices. Each element in the resulting tensor $\vec{C}$ is simply the product of the elements with the corresponding indices from $\vec{A}$ and $\vec{B}$.

$$\vec{C} = \vec{A} * \vec{B} \quad \text{where} \quad c_{i_1 \cdots i_n} = a_{i_1 \cdots i_n} \times b_{i_1 \cdots i_n}$$

**Outer Product**  The outer product of two tensors $\vec{A}$ and $\vec{B}$ having order $m$ and $n$ respectively is a tensor $\vec{C}$ of order $m+n$. Each element of $\vec{C}$ is the product of the elements of $\vec{A}$ and $\vec{B}$ whose aggregate indices correspond to its own indices.

$$\vec{C} = \vec{A} \circ \vec{B} \quad \text{where} \quad c_{i_1 \cdots i_m j_1 \cdots j_n} = a_{i_1 \cdots i_m} \times b_{j_1 \cdots j_n}$$

**Inner Product**  The inner product of two tensors $\vec{A}$ and $\vec{B}$ is a tensor formed by taking the outer product of $\vec{A}$ and $\vec{B}$ and then summing up over common indices that appear both in $\vec{A}$ and $\vec{B}$. If $\vec{A}$ is of order $m$, $\vec{B}$ is of order $n$ and they have $k$ common indices then the inner product $\vec{C}$ is a tensor of order $(m-k)+(n-k)$.

$$\vec{C} = \vec{A} \bullet \vec{B} \quad \text{where} \quad c_{i_1 \cdots i_{m-k} j_1 \cdots j_{n-k}} = \sum_{l_1, \cdots, l_k} a_{i_1 \cdots i_{m-k} l_1 \cdots l_k} \times b_{l_1 \cdots l_k j_1 \cdots j_{n-k}}$$

In order to see how these standard tensor operations are applicable to the belief revision procedure, we consider a few relevant examples.

Let $\vec{\lambda}^*_{Y_j}$ and $\vec{\pi}^*_{U_i}$ be vectors (or, equivalently, tensors of order 1) whose elements are the messages a node $X$ receives from its children and its parents respectively:

$$\vec{\lambda}^*_{Y_j} = \left[ \lambda^*_{Y_j}(x_1), \cdots, \lambda^*_{Y_j}(x_r) \right] \quad \text{where } r \text{ is the number of possible values for } X$$

$$\vec{\pi}^*_{U_i} = \left[ \pi^*_X(u_{i_1}), \cdots, \pi^*_X(u_{i_{r(i)}}) \right] \quad \text{where } r(i) \text{ is the number of possible values for } U_i$$

The term product of all $\vec{\lambda}^*_{Y_j}$ vectors is another vector $\vec{\Lambda}^*$ of length $r$ given by

$$\vec{\Lambda}^* = \vec{\lambda}^*_{Y_1} * \cdots * \vec{\lambda}^*_{Y_m} = \left[ \prod_{j=1}^{m} \lambda^*_{Y_j}(x_1), \cdots, \prod_{j=1}^{m} \lambda^*_{Y_j}(x_r) \right]$$

The outer product of all $\vec{\pi}^*_{U_i}$ vectors is a tensor $\vec{\Pi}^*$ of order $n$ given by

$$\vec{\Pi}^* = \vec{\pi}^*_{U_1} \circ \cdots \circ \vec{\pi}^*_{U_n} \quad \text{where} \quad \pi^*_{k_1 \cdots k_n} = \prod_{i=1}^{n} \pi^*_X(u_{i_{k_i}})$$

We can consider the set of fixed probabilities $P(x \mid u_1, \cdots, u_n)$ as elements of a tensor $\vec{P}$ of order $n+1$. Now if we compute the inner product of $\vec{P}$ with $\vec{\Pi}^*$ we obtain a tensor of order 1 (the indices for the $U_i$ are common to both tensors):

$$\vec{P} \bullet \vec{\Pi}^* = \left[ \sum_{i_1, \cdots, i_n} P(x_1 \mid u_{i_1}, \cdots, u_{i_n}) \prod_{k=1}^n \pi_X^*(u_{i_k}), \cdots, \sum_{i_1, \cdots, i_n} P(x_r \mid u_{i_1}, \cdots, u_{i_n}) \prod_{k=1}^n \pi_X^*(u_{i_k}) \right]$$

The elements of this last inner product are very similar to the formula used to compute $BEL^*(x)$. To see this, we can rewrite the $BEL^*(x)$ equation as follows:

$$BEL^*(x) = \alpha \prod_{j=1}^m \lambda_{Y_j}^*(x) \max_{u_k, 1 \le k \le n} P(x \mid u_1, \cdots, u_n) \prod_{i=1}^n \pi_X^*(u_i)$$

The difference between the last portion of this term and the elements in the above inner product is that the $BEL^*(x)$ computation requires us to maximize over all elements $u_k$ rather than taking a sum.

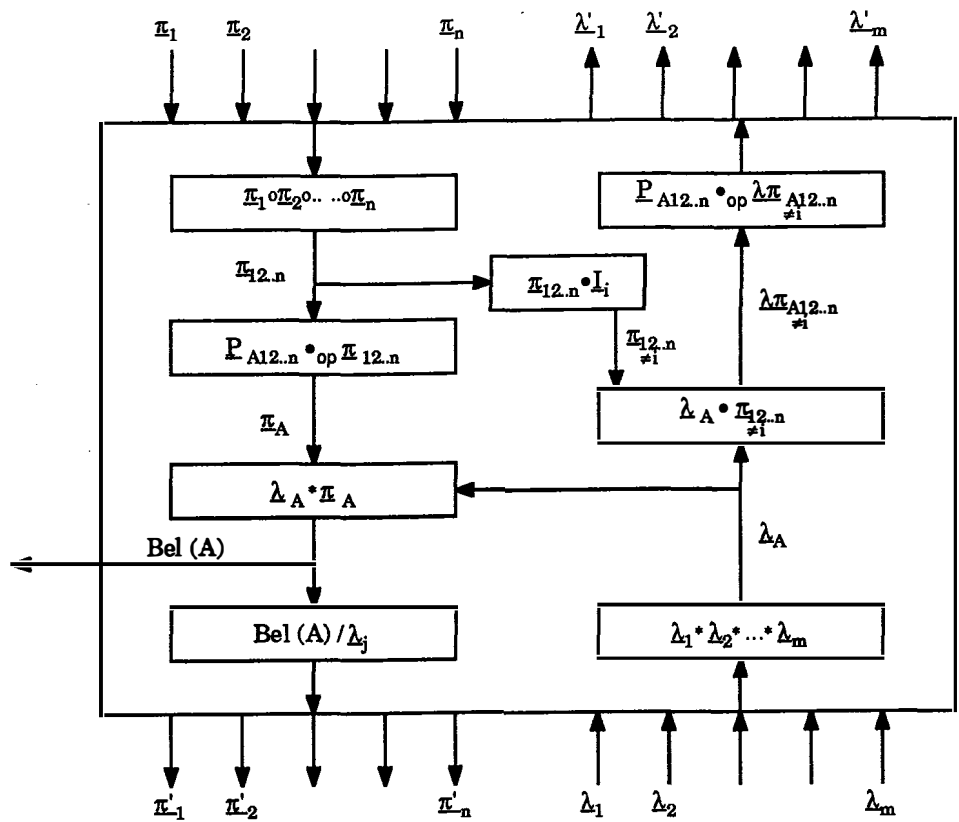

Figure 1. Combined updating of belief and belief commitment (the operator $\bullet_{op}$ represents the standard or modified inner product depending on whether op is + or max). In belief commitment all the $\pi$s, $\lambda$s, and Bels should be changed to $\pi^*$s, $\lambda^*$s, and Bel*s.

But we are free to redefine our inner product operator to do the same thing. That is, we can simply define the elements of the inner product tensor to be given by

$$c_{i_1 \cdots i_{m-k} j_1 \cdots j_{n-k}} = \max_{l_1, \cdots, l_k} a_{i_1 \cdots i_{m-k} l_1 \cdots l_k} \times b_{l_1 \cdots l_k j_1 \cdots j_{n-k}}$$

Now if we denote this new inner product with the symbol $\bullet_{max}$, then the $BEL^*(x)$ computation can be written in tensor notation as

$$\vec{BEL}^* = \alpha \vec{\Lambda}^* \ast (\vec{P} \bullet_{max} \vec{\Pi}^*)$$

353

Moreover, it is clear that we can use similar methods to compute the messages that node $X$ will send to its neighbors. The vector $\vec{\pi}^*_{Y_j}$ destined for child $Y_j$ can be computed by term-by-term division of the elements of $\vec{BEL}^*$ by the elements of $\vec{\lambda}^*_{Y_j}$. The vector $\vec{\lambda}^*_X$ destined for parent $U_i$ can be computed just like $\vec{BEL}^*$ except that we replace the vector $\vec{\pi}^*_{U_i}$ with a unit vector $(1, \cdots, 1)$ of equal length when computing the outer product $\vec{\Pi}^*$.

These tensor computations turn out to be similar to the method proposed by Kim [6] for belief updating, which requires the standard inner product using summation. This means that beliefs and belief commitments can be computed in one uniform scheme as shown in Figure 1.

**Conclusion**

The tensor product method of revising belief commitments uses enumeration to solve what would otherwise be some very difficult optimization problems. This makes Pearl's [11] belief revision computation practical even when the set of link probabilities does not fit any well understood, canonical model of causal interactions. Since the classification problems of interest here involve networks with probabilities that are difficult to characterize analytically, it was important to devise a version of belief revision that is truly general. All that the tensor product method requires is the set of link probabilities, which is given as part of the specification of every network.